\documentclass[a4paper, 10pt, conference]{ieeeconf}      
\usepackage{FG2026}

\FGfinalcopy 

\IEEEoverridecommandlockouts                              
\usepackage{graphics} 
\usepackage{epsfig} 
\usepackage{mathptmx} 
\usepackage{times} 
\usepackage{amsmath} 
\usepackage{amssymb}  
\usepackage{multirow}
\usepackage{cuted}
\usepackage{graphicx}  
\usepackage{booktabs}
\usepackage{subcaption}
\usepackage{makecell}
\usepackage[inkscapelatex=false]{svg} 
\usepackage{tikz}
\usetikzlibrary{positioning}
\usetikzlibrary{calc}
\def\FGPaperID{186

} 

\title{\LARGE \bf
FD-MAD: Frequency-Domain Residual Analysis for Face Morphing Attack Detection
}

\author{\parbox{16cm}{\centering
    {\large Diogo J. Paulo$^{1,2,3}$, Hugo Proença$^{1,3}$ and João C. Neves$^{1,2}$}\\
    {\normalsize
    $^1$ University of Beira Interior, Portugal \quad
    $^2$ NOVA LINCS \quad
    $^3$ IT: Instituto de Telecomunicações}}
}

\begin{document}

\ifFGfinal
\thispagestyle{empty}
\pagestyle{empty}
\else
\author{Anonymous FG2026 submission\\ Paper ID \FGPaperID \\}
\pagestyle{plain}
\fi
\maketitle

\begin{abstract}
Face morphing attacks present a significant threat to face recognition systems used in electronic identity enrolment and border control, particularly in single-image morphing attack detection (S-MAD) scenarios where no trusted reference is available. In spite of the vast amount of research on this problem, morph detection systems struggle in cross-dataset scenarios. To address this problem, we introduce a region-aware frequency-based morph detection strategy that drastically improves over strong baseline methods in challenging cross-dataset and cross-morph settings using a lightweight approach. Having observed the separability of bona fide and morph samples in the frequency domain of different facial parts, our approach 1) introduces the concept of residual frequency domain, where the frequency of the signal is decoupled from the natural spectral decay to easily discriminate between morph and bona fide data; 2) additionally, we reason in a global and local manner by combining the evidence from different facial regions in a Markov Random Field, which infers a globally consistent decision. The proposed method, trained exclusively on the synthetic morphing attack detection development dataset (SMDD), is evaluated in challenging cross-dataset and cross-morph settings on FRLL-Morph and MAD22 sets. Our approach achieves an average equal error rate (EER) of 1.85\% on FRLL-Morph and ranks second on MAD22 with an average EER of 6.12\%, while also obtaining a good bona fide presentation classification error rate (BPCER) at a low attack presentation classification error rate (APCER) using only spectral features. These findings indicate that Fourier-domain residual modeling with structured regional fusion offers a competitive alternative to deep S-MAD architectures.
\end{abstract}    
\section{Introduction}
\label{sec:intro}
Face morphing attacks represent a significant threat to face recognition systems used in electronic identification (eID) enrolment and border control. When the facial images of two individuals are blended into a single plausible face, malicious actors can obtain documents that are accepted under both identities. These attacks exploit the tolerance of modern recognition systems to intra-class variation and have been demonstrated through classical landmark-based warping~\cite{landmarkbased,frll_morph}, GAN-based synthesis~\cite{stylegan_morph,damer2018morgan,zhang2021mipgan}, and, more recently, diffusion models~\cite{mordiff,blasingame2024leveraging,fastdim,greedydim,zhang2024diffmorpher}. This threat has driven extensive research on single-image morphing attack detection (S-MAD), where the goal is to determine whether a given face image is bona fide or morphed, in line with emerging standards such as ISO/IEC~30107-3~\cite{ISO}.
\begin{figure}[th]
    \centering
    \includegraphics[width=.99\linewidth]{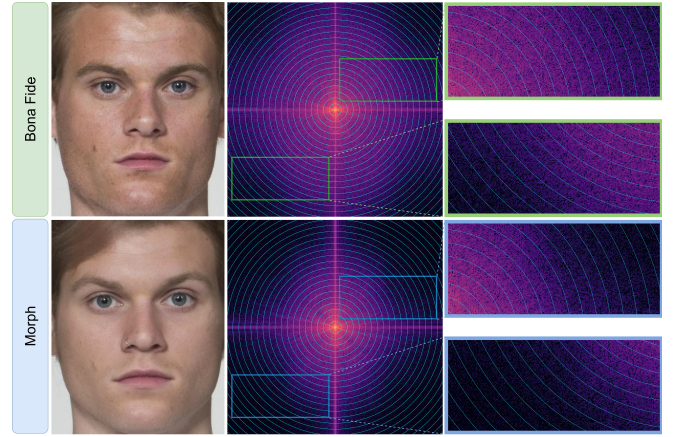}
    \caption{\textbf{Comparison between frequency-domain data of a bona fide face and a morph sample.} The key observation is that the morphs typically differ from the bona fide samples in the mid-high frequencies, which is used as the insight for the proposed method.} 
    \label{fig:teaser}
    \vspace{-0.5em}
\end{figure}
The earliest S-MAD approaches used handcrafted texture and color features in combination with classical classifiers~\cite{texture_morphs,ramachandra2019detecting}. More recent research~\cite{smdd_inception_mixfacenet, synmad22, orthomad, idistill} has focused on developing deep learning architectures trained on extensive datasets of bona fide and morphed images. Additionally, transformer-based and foundational models have been used, including ~\cite{transformermad} and ~\cite{caldeira2025madation}. Despite substantial progress, these methods often require complex architectures, which increase computational demands and complicate deployment in large-scale and time-sensitive environments. More recent advances include unsupervised and self-supervised S-MAD methods, which reduce the reliance on labelled data and aim to improve robustness to unseen attacks by modelling bona fide data distributions or leveraging self-supervised pretext tasks~\cite{selfmad,selfmad++,huber2025fx}.  However, achieving strong cross-morph and cross-dataset generalization remains particularly challenging. Models trained on a given morphing pipeline often suffer performance drops when evaluated on unseen morphing attacks. Recent work has begun to exploit frequency-domain information and frequency-aware data augmentation to address this limitation. For example, Huber \textit{et al.}~\cite{huber2025fx} operate in the differential morphing attack detection (D-MAD) setting, while other approaches mainly regularize deep learning models by creating synthetic frequency perturbations~\cite{selfmad++}. Overall, frequency-domain cues have so far received little attention as primary features for S-MAD, despite their potential to capture global and local spectral regularities in the Fourier domain. Hence, we propose a residual frequency-domain S-MAD framework that directly targets cross-morph generalization by obtaining invariant global–local features in the Fourier domain, without relying on heavy architectures prone to overfitting and to capturing dataset or morph-specific biases.

In this work, we introduce a global–local framework that explicitly models deviations from the natural power-law spectral decay. Our design is motivated by extensive evidence that natural images exhibit characteristic magnitude spectra that approximately follow a $f^{-\alpha}$ law~\cite{naturalpattern1,naturalpattern2,naturalpattern3}, while synthetic and manipulated imagery often deviates from this behaviour~\cite{dzanic2020fourier,chandrasegaran2021closer,corvi2023intriguing,tan2024frequency}. Based on these observations, we represent each face image in a frequency domain residual space that highlights deviations from the natural spectral trend (Figure~\ref{fig:teaser}). At the global level, we obtain a radial log-magnitude spectrum, estimate an image-specific spectral baseline, and treat the difference as a compact residual descriptor, which is then fed to a classifier. At the local level, we apply the same residual construction to four semantic regions (left and right eye, nose, and mouth), obtaining region-wise descriptors. To capture structured dependencies between these regions, such as local artefacts affecting only a subset of them, we use a pairwise Markov Random Field (MRF) over the region labels, to encourage coherent decisions across neighboring regions and avoid scattered or inconsistent labels. This approach provides a probabilistic global–local morphing score that is both computationally efficient and interpretable.

We evaluate the proposed method in a challenging cross-dataset and cross-morph scenario. Models are trained on privacy-friendly synthetic data from SMDD~\cite{smdd_inception_mixfacenet} and tested on FRLL-Morph~\cite{frll_morph} and MAD22~\cite{synmad22}, which include a broad range of landmark-based and GAN-based attacks~\cite{zhang2021mipgan,stylegan,stylegan_morph,qin2020low}. Our approach consistently outperforms or matches several recent supervised methods on FRLL-Morph~\cite{frll_morph}, and achieves competitive results on MAD22~\cite{synmad22} while remaining significantly more lightweight than transformer-based and multi-task architectures. Ablation studies demonstrate that global and local Fourier residuals provide complementary information and that MRF-based fusion is essential for addressing both classical and GAN-based morphs. t-SNE visualizations further reveal well-structured, attack-sensitive embeddings in the frequency domain, reinforcing the interpretability of our approach.

In summary, our main contributions are as follows:
\begin{itemize}

    \item A novel S-MAD framework that operates entirely in the Fourier domain, modelling deviations from the natural $f^{-\alpha}$ spectral law using compact global and local residual features, that amplify morphing artefacts.
    \item An MRF-based fusion of semantic facial regions, which captures structured dependencies between local decisions and improves robustness to heterogeneous, region-specific morphing artefacts, including those from GAN-based attacks.
    \item Extensive cross-morph experiments on FRLL-Morph~\cite{frll_morph} and MAD22~\cite{synmad22}, demonstrating that our approach achieves state-of-the-art or highly competitive performance compared to recent supervised S-MAD methods, while remaining computationally efficient, making it suitable for real-world deployment.
\end{itemize}
The code will be made publicly available.

\section{Related Work}
\label{sec:related_work}
\noindent\textbf{Face Morphing.}
Previous face morphing techniques demonstrated that blending the geometry and texture of two identities can produce a synthetic face image that matches both contributors in automatic face verification systems. \cite{ferrara} was among the first to show the vulnerability of passport-style biometric systems to such attacks, motivating substantial research into morphing generation and detection. Subsequent works improved the realism and controllability of morphs, ranging from landmark-based image warping~\cite{landmarkbased} to more advanced methods which interpolate the latent representations like GAN-based morphs~\cite{zhang2021mipgan, colbois2023approximating} and more recently diffusion-based morphs~\cite{fastdim, greedydim, zhang2024diffmorpher, mordiff, blasingame2024leveraging}. 

\vspace{0.5em}
\noindent\textbf{Morphing Attack Detection.}
Morphing Attack Detection (MAD) is typically categorized into single-image MAD (S-MAD) and differential MAD (D-MAD). D-MAD, which compares a live image with a trusted reference, achieves strong accuracy but is less relevant for passport issuance scenarios where no live comparison is available. S-MAD is therefore a more challenging and widely studied setting. Hence, several studies have developed S-MAD systems~\cite{idistill, ramachandra2019detecting, orthomad, pwmad, transformermad, texture_morphs, caldeira2025madation, synmad22, selfmad, selfmad++}, in order to detect whether the investigated image is a morph based only on its features.

Early S-MAD approaches relied on handcrafted texture descriptors~\cite{texture_morphs} such as LBP, BS1F, or S1FT histograms, exploiting inconsistencies in facial regions introduced by the morphing warp. However, as morphing attacks became more diverse, these features lacked robustness against unseen morphing methods. Deep learning–based S-MAD approaches have significantly improved performance on controlled benchmarks; however, they exhibit limited robustness under cross-morphing scenarios. Although approaches such as IDistill~\cite{idistill} or MixFaceNet-MAD~\cite{smdd_inception_mixfacenet} achieve high accuracy on SMDD, they degrade significantly on FRLL-morphs such as AMSL, indicating a strong dependence on the texture statistics of the training dataset. Even MADation~\cite{caldeira2025madation}, which uses CLIP’s large-scale pretraining and lightweight LoRA~\cite{lora} fine-tuning, illustrates these limitations. Even though MADation~\cite{caldeira2025madation} achieves good performance in a large portion of the MAD22 and MorDIFF evaluation settings, the authors report notable performance drops for some unseen morphing methods (e.g., MIPGAN-I/II when trained on SMDD). This reinforces that even foundation models used in S-MAD methods remain sensitive to cross-morph scenarios. These findings evidence that current deep learning S-MAD systems, while powerful, are inherently vulnerable to dataset bias, texture bias overfitting, and poor cross-morph generalization, especially when the training data come from a single dataset such as SMDD~\cite{smdd_inception_mixfacenet}.

In~\cite{pca_morph}, the authors proposed analyzing texture artifacts using PCA projections, highlighting the utility of interpretable, low-dimensional texture representations rather than features derived from black-box models. However, this approach relies only on global image descriptors. It does not model region-specific inconsistencies or dependencies among facial areas—artifacts that are often amplified by the morphing process. Hence, our method extends this family by combining global descriptors with local spectral cues and structured probabilistic reasoning.

\vspace{0.5em}
\noindent\textbf{Frequency Artifacts in Generated Morphs.}
Extensive research in image forensics demonstrates that generative models and image warping techniques leave characteristic traces in the frequency domain~\cite{dzanic2020fourier, corvi2023intriguing, chandrasegaran2021closer, tan2024frequency}. Furthermore, these frequency irregularities are not limited to deepfakes or GAN/diffusion synthesis; face morphing operations also distort natural spectral statistics, and these irregularities have been studied in some prior works~\cite{huber2025fx, tapia_fourier, wb_avcivas}. It is worth mentioning that \cite{huber2025fx} studies frequency cues in the D-MAD setting, which assumes access to a trusted reference image; therefore, it is not directly comparable to our S-MAD scenario and is cited here primarily to support the relevance of frequency domain artifacts. Face morphing combines geometric warping with pixel-level blending, both of which disrupt local textures, smooth natural edges, and attenuate high-frequency content—particularly around manipulated regions such as eyes, nose, or mouth. Recent works in morphing forensics further demonstrate that, because morphs introduce artifacts at high frequencies, the altered images deviate from the natural $f^{-\alpha}$ power-law behavior~\cite{naturalpattern1, naturalpattern2, naturalpattern3}. These observations motivate the use of frequency domain features, which can generalize more robustly across morphing pipelines than spatial CNN features, which are prone to overfitting dataset-specific textures.
\section{Methodology}
\label{sec:proposed_method}
\begin{figure*}[ht]
    \centering
    \includegraphics[width=\linewidth]{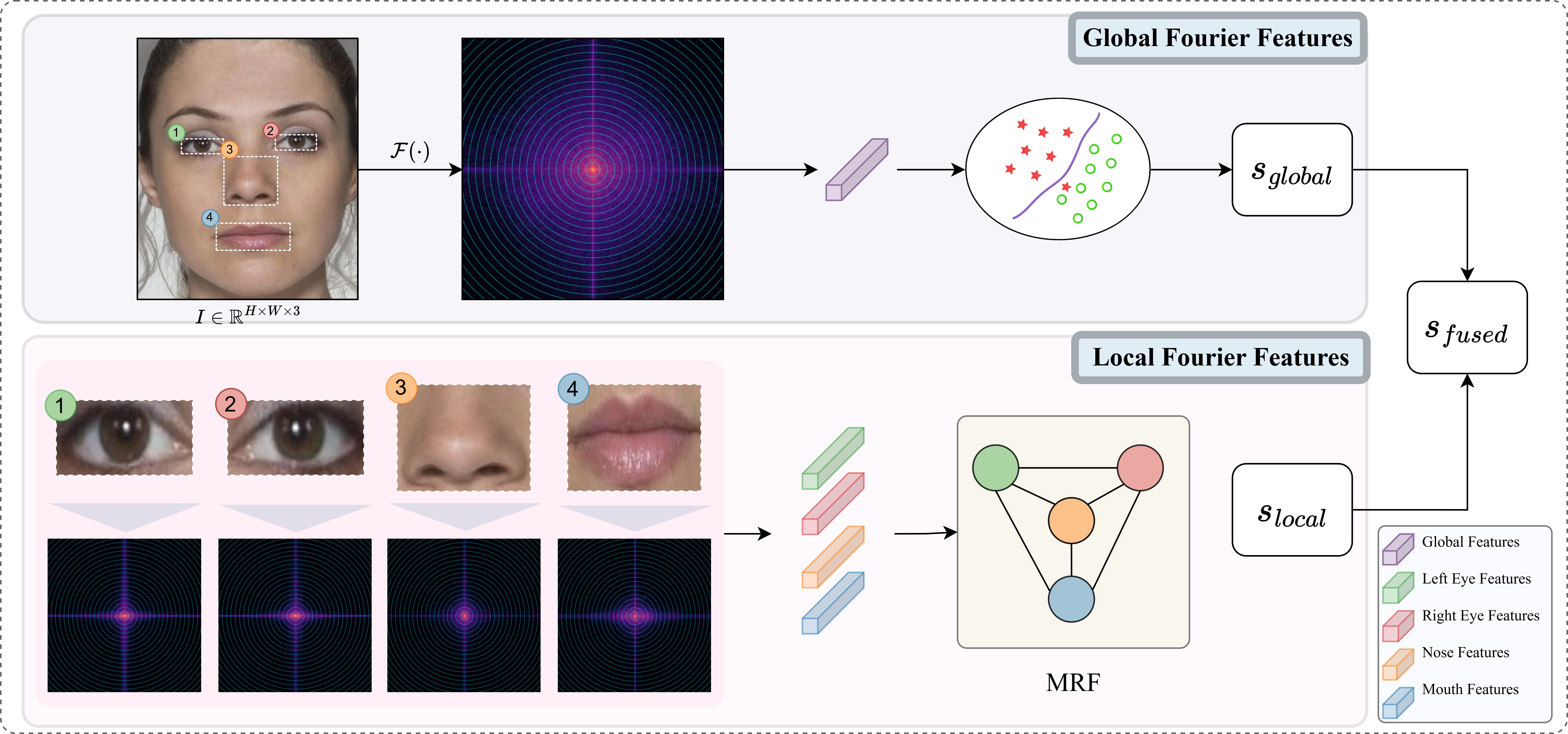}
    \caption{\textbf{Overview of the proposed Fourier‐domain morphing detector.} The input face is transformed to the frequency domain and summarized into (top) global azimuthally averaged FFT magnitude profiles used by an SVM classifier to obtain a global score ($s_{global}$), and (bottom) local profiles extracted from the left eye, right eye, nose, and mouth, whose region-wise classifiers are fused by a Markov Random Field (MRF) to produce a local score ($s_{local}$). The global and local scores are then combined through a weighted fusion to obtain the final morphing detection score.}
    \label{fig:proposed_method}
\end{figure*}
In order to enhance the detection of face morphing attacks, we propose a lightweight and interpretable approach based on the Fourier frequency domain. Our method is based on the observation that natural images follow a predictable spectral decay, typically following a $f^{-\alpha}$ distribution of magnitude across spatial frequencies~\cite{naturalpattern1, naturalpattern2, naturalpattern3}, where $\alpha$ is the spectral decay exponent (i.e., the slope of the radial spectrum in log-log space), which varies with image content but is typically stable for natural images. In contrast, face morphing attacks tend to distort this natural behavior by introducing artifacts and irregularities in the frequency spectrum, particularly in the mid and high-frequency bands. We obtain both global and local (region-wise) residual Fourier magnitude features, removing the estimated natural $f^{-\alpha}$ spectral trend to emphasize morphing artifacts, and train a classifier on the resulting residuals. To integrate region-specific evidence while capturing structured region-level dependencies, we introduce a Markov Random Field (MRF) model whose unary potentials derive from local classifiers and whose pairwise potentials penalize label disagreements, encouraging globally consistent region labels while still allowing the unary terms to capture localized morphing artifacts. The final morphing score is obtained by aggregating the global classifier output and the MRF local decisions within a unified decision function, thereby using complementary frequency cues. An overview of the proposed framework is illustrated in Figure~\ref{fig:proposed_method}.

\subsection{Global Fourier Residual Features}
\label{subsec:global}

Given an RGB image $I \in \mathbb{R}^{H \times W \times 3}$ we obtain, for each channel, the two–dimensional discrete Fourier transform (DFT):
\begin{equation}
    F(u,v) = \mathcal{F}\{ I(x,y) \},
\end{equation}
where $\mathcal{F}(\cdot)$ denotes the 2D DFT mapping the spatial signal $I(x,y)$ to frequency coordinates $(u,v)$. We then compute its centered log-magnitude spectrum as follows:
\begin{equation}
    S(u,v) = \log\!\left(1 + \big| F(u,v) \big| \right).
\end{equation}
To obtain a compact and rotation–invariant representation, we split the magnitude spectrum into $K$ concentric radial frequency bands. Let $R_k$ denote the set of frequency coordinates in the $k$-th ring.
We calculate the azimuthally–averaged band energy as:
\begin{equation}
    b_k = \frac{1}{|R_k|} \sum_{(u,v)\in R_k} S(u,v),
\end{equation}
forming a 1D global spectral profile
$b = [b_1, \dots, b_K]$. The number of radial frequency bands $K$ is determined by the maximum attainable radius in the Fourier domain, that is, $K = \frac{N}{2}\sqrt{2}$, for an image of size $N \times N$. Additionally, we decided to keep the entire spectrum (no low or high-frequency bands are discarded) to avoid discarding potentially discriminative frequency content.

\begin{figure}[ht]
    \centering
    \includegraphics[width=\linewidth]{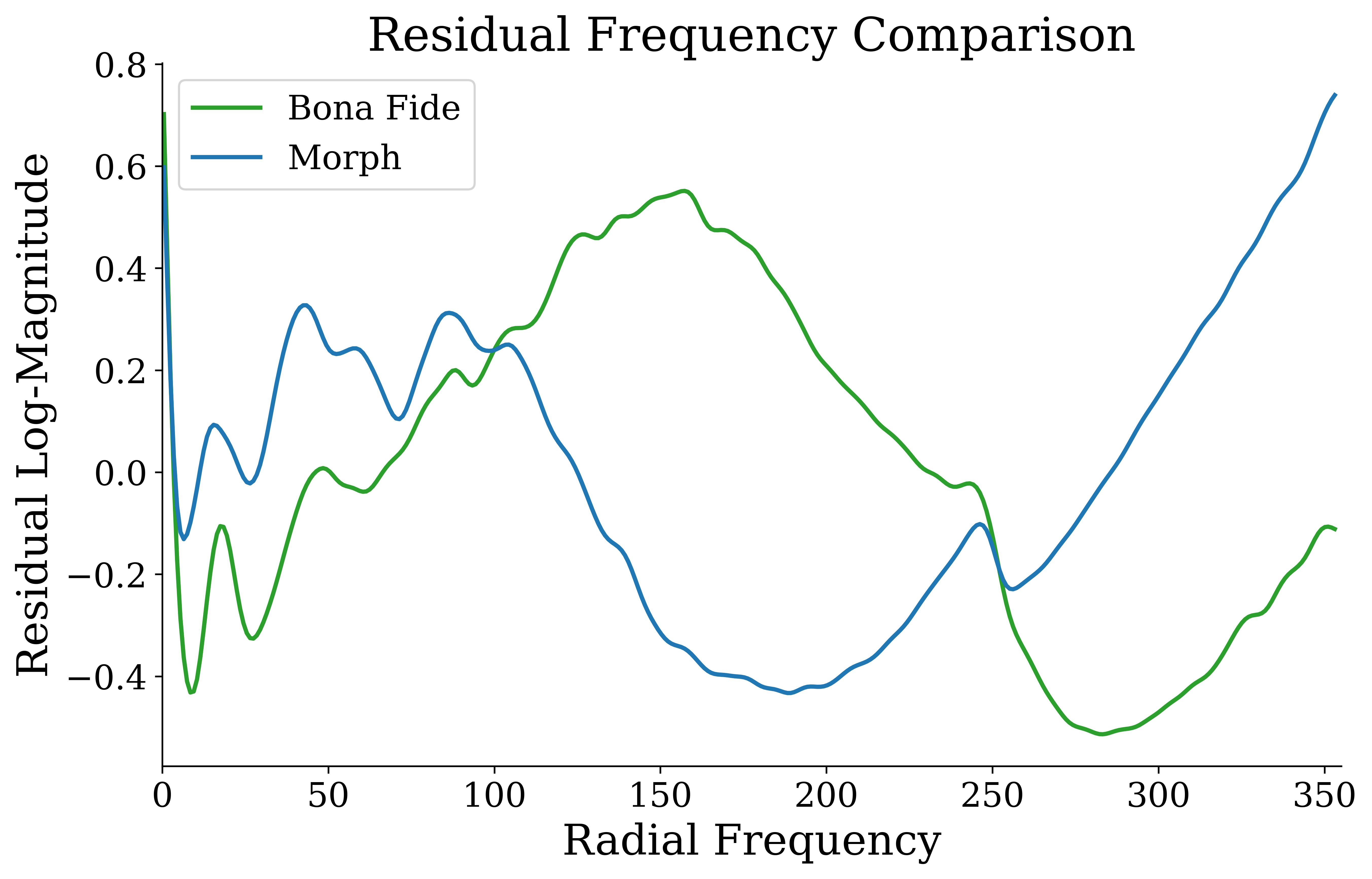}
    \caption{\textbf{Residual radial-frequency profiles of a bona fide image and its corresponding morph.} Compared to the bona fide image, the morph exhibits systematic deviations, particularly in the mid-to-high frequency bands, consistent with blending and interpolation artifacts introduced during morphing.}
    \label{fig:residuals}
\end{figure}

\vspace{0.5em}
\noindent\textbf{Power-Law Spectral Baseline Removal.}
Natural images approximately follow a power–law spectral decay 
$S(f) \propto f^{-\alpha}$, which is linear in log–log space.
We therefore fit a line to the radial log-magnitude profile in log-frequency space:
\begin{equation}
    \log S(f) = a + b \log f,
\end{equation}
where $a$ and $b$ are the least squares intercept and slope of the fitted power-law baseline. That is, under the assumption $S(f)=C f^{-\alpha}$, we have $a=\log C$ and $b=-\alpha$. 
We define $f \in \{f_1,\ldots,f_K\}$ as the discrete radial frequencies associated with the $K$ concentric rings. This yields the estimated log baseline $\widehat{\log S}(f)= a + b \log f$.
The residuals are then obtained as:
\begin{equation}
    r(f) = \log S(f) - \widehat{\log S}(f).
\end{equation}
In the residual space, morphs are expected to distinguish from bona fide images, due to the irregularities of synthetic images in the mid and high frequencies (Figure~\ref{fig:residuals}).

To further remove redundancy, the RGB residual profiles are concatenated, standardized, and reduced using PCA, obtaining $x_{r}$. Subsequently, these features are then provided to a classifier to distinguish between bona fide and morphs, which outputs a global score $s_{\mathrm{global}} = p(z_r = 1 \mid x_{\mathrm{r}})$, where $z_r \in \{0,1\}$ denotes morph ($0$) or bona fide ($1$).

\subsection{Local Region-Wise Spectral Descriptors}
\label{subsec:local_features}
Global Fourier residual features mostly capture global trend inconsistencies, while morphing artifacts often appear in localized and heterogeneous regions. For instance, warping boundaries and interpolation errors introduce fine-grained irregularities around semantic regions such as the eyes, nose, and mouth. Thus, to complement the global representation and target localized distortions, we extract region-wise Fourier residual features from facial subregions, yielding more subtle morphing cues that may be hidden or averaged in the global features.

For this, a face landmark detector is used to localize key facial areas and crop semantic region patches. 
The resulting patches are resized to a fixed resolution and processed with the same pipeline as in Section~\ref{subsec:global} to obtain a region-specific Fourier feature vector $x_r$.

A logistic regression $C_r$ is then trained per region, yielding posterior probabilities:
\begin{equation}
    p_r(z_r = 1 \mid x_r).
\end{equation}
We convert these into unary potential terms,
\begin{equation}
    \psi_r(z_r) = -\log p_r(z_r \mid x_r).
\end{equation}

These unary potentials form the basis of our structured fusion model, serving as input to the MRF; they are obtained from the region-wise classifiers. We chose to use the log-domain instead of raw probabilities for two reasons. First, multiplying many small probabilities can lead to numerical underflow, whereas log-space converts these products into simple sums that remain well-conditioned. Second, the MRF energy function is additive, and log-potentials allow the unary and pairwise terms to be combined in a coherent linear form, enabling exact inference and an appropriately weighted contribution from each region. Thus, the log-domain formulation ensures both computational stability and a principled probabilistic interpretation within the structured fusion model.

\subsection{Structured Fusion with Markov Random Field}
\label{subsec:mrf}
We use a pairwise MRF to obtain a consensus across different region estimates, which would not be possible by simply averaging local predictions. This would assume conditional independence across regions. In practice, bona fide faces tend to produce globally consistent decisions across regions, whereas morphs often exhibit heterogeneous, region-level cues: for example, local blending artifacts may strongly affect the mouth and one eye while leaving other areas visually plausible. To capture this behavior, we use pairwise potentials that penalize label disagreements, encouraging globally consistent region labels while still allowing the unary terms to capture local morphing artifacts. As a result, the MRF serves as a context-aware fusion mechanism that exploits dependencies between regions rather than treating them independently.

\vspace{0.5em}
\noindent\textbf{MRF Formulation.} Let $z = (z_1, \dots, z_R)$ denote the unknown binary labels of $R$ facial regions.
We build a fully connected MRF with a potential function defined as:
\begin{equation}
    E(z) = 
    \sum_{r=1}^{R} \psi_r(z_r)
    + \sum_{(i,j)\in \mathcal{E}} \psi_{ij}(z_i,z_j),
\end{equation}
where $\mathcal{E}$ denotes the set of region pairs, $\mathcal{E}=\{(i,j)\mid 1\le i<j\le R\}$.

The unary potentials $\psi_r(z_r)$ are defined in Sec.~\ref {subsec:local_features} and reflect the region–specific spectral evidence. Additionally, to model structured inconsistencies between facial regions, we also employ an Ising pairwise term. For each pair of connected regions $(i,j)$, the pairwise potential is defined as
\begin{equation}
\psi_{ij}(z_i, z_j) = \beta |z_i - z_j|, \quad \beta > 0,
\label{eq:pairwise}
\end{equation}

where $\beta$ controls the strength of the agreement constraint between regions. This pairwise potential penalizes label disagreements and thus encourages smooth, globally consistent labelings across regions. This allows the unary terms to express local morphing evidence while the MRF regularizes overly fragmented decisions.

\vspace{0.5em}
\noindent\textbf{Exact Inference.} Since the number of regions $R$ is limited when analyzing human faces, exact inference is tractable. Hence, we enumerate all $2^R$ configurations and calculate
\begin{equation}
    P(z \mid x)
    = \frac{\exp\!\big(-E(z)\big)}
           {\displaystyle \sum_{z' \in \{0,1\}^R} \exp\!\big(-E(z')\big)}.
\end{equation}
\begin{figure}[ht]
    \centering
    \includegraphics[width=\linewidth]{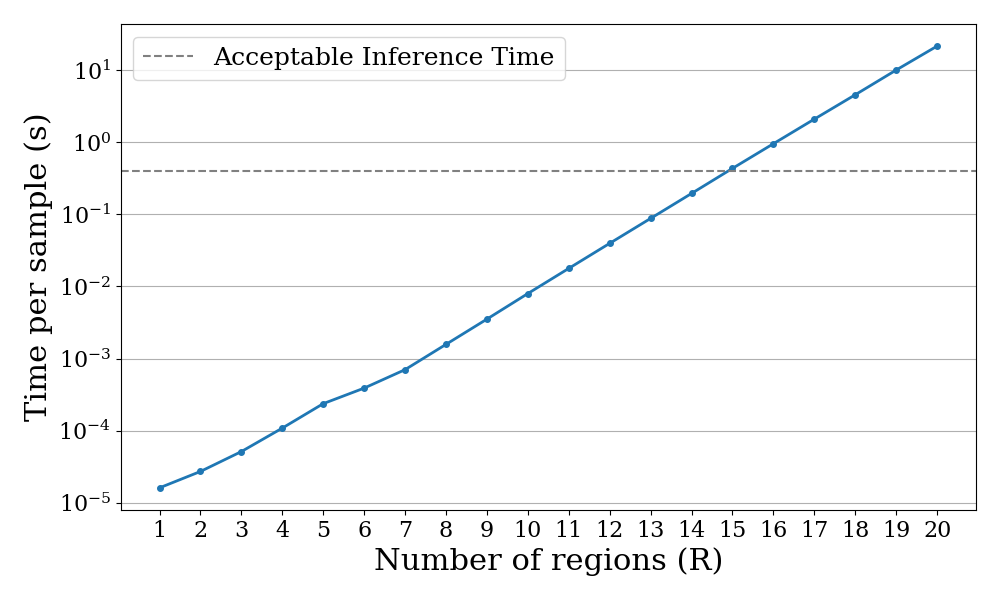}
    \caption{\textbf{Exact MRF inference time versus number of facial regions $R$.} While the complexity scales exponentially with $R$ due to full enumeration, faces admit only a limited number of meaningful regions, and inference remains tractable even for conservative upper bounds on $R$.}
    \label{fig:inference_time}
\end{figure}

As illustrated in Figure~\ref{fig:inference_time}, although the resulting inference time grows exponentially with $R$, this does not pose a practical limitation for face morphing detection, since faces admit only a small number of meaningful semantic regions and region-based methods, including ours, operate in this setting.
Finally, we output the expected fraction of bona fide labels:
\begin{equation}
    s_{\mathrm{local}}
    = \mathbb{E}_{P(z \mid x)}\!\left[ \frac{1}{R} \sum_{r=1}^{R} z_r \right]
    = \sum_{z \in \{0,1\}^R}
      \left(\frac{1}{R} \sum_{r=1}^{R} z_r \right)
      P(z \mid x).
\end{equation}
This constitutes the local morphing detection score.

\subsection{Global-Local Fusion}
\label{subsec:fusion}

The global score $s_{\mathrm{global}}$ (holistic spectral deviations)
and the MRF score $s_{\mathrm{local}}$ (structured region-level evidence)
provide complementary information. Therefore, we combine them using a weighted fusion:
\begin{equation}
    s_{\mathrm{fused}} 
    = \lambda \, s_{\mathrm{global}}
    + (1-\lambda) \, s_{\mathrm{local}},
\end{equation}
where $\lambda$ is tuned on the SMDD validation set.
This fusion leverages both global robustness and region–specific structural coherence, yielding improved cross–morph generalization.

\section{Experimental Setup}
\label{sec:databases}
To assess the generalization capability of the proposed method, we rely on three morphing datasets, SMDD~\cite{smdd_inception_mixfacenet}, FRLL-Morph~\cite{frll_morph}, and MAD22~\cite{synmad22}.
SMDD is a synthetic StyleGAN2-based development dataset with OpenCV morphs, while FRLL-Morph and MAD22 are FRLL-based benchmarks containing landmark and GAN-based morphs and are identity-disjoint from SMDD. Furthermore, we use MorDIFF~\cite{mordiff}, an extension of MAD22, which provides additional FRLL-based diffusion morphs, enabling evaluation under stronger and more realistic attacks.

\subsection{Evaluation Metrics}
\label{subsec:metrics}

All experiments follow the ISO/IEC~30107-3~\cite{ISO} standard for biometric presentation attack detection. We report the Attack Presentation Classification Error Rate (APCER), defined as the proportion of morphing attacks incorrectly classified as bona fide, and the Bona Fide Presentation Classification Error Rate (BPCER), defined as the proportion of bona fide presentations incorrectly classified as attacks. We also report the Equal Error Rate (EER), corresponding to the operating point at which APCER equals BPCER. In line with the ISO/IEC~30107-3~\cite{ISO} recommendations and prior S-MAD work~\cite{pwmad, idistill, orthomad}, we additionally report BPCER at fixed APCER thresholds of 1\% and 20\%. Reporting BPCER@APCER is particularly informative from a security perspective, as it quantifies the false rejection rate under a controlled false acceptance rate. Since minimizing APCER is critical to prevent morphing attacks from being falsely accepted, BPCER@APCER directly reflects the cost of enforcing a given security level.

\subsection{Evaluation Protocols}
\label{subsec:protocols}

The MAD22 benchmark~\cite{synmad22} and SMDD~\cite{smdd_inception_mixfacenet} provide standardized experimental setups for single morphing attack detection, which include multiple morphing tools. Building on these protocols, we adopt standard cross-dataset and cross-morph settings commonly used in previous S-MAD studies~\cite{pwmad, idistill, orthomad, smdd_inception_mixfacenet, caldeira2025madation}. For the first protocol, we train on SMDD~\cite{smdd_inception_mixfacenet} and test on FRLL-Morph~\cite{frll_morph}, which yields a combined cross-dataset and cross-morph scenario (OpenCV-based synthetic training vs.\ multiple morphing tools on FRLL). For the second protocol, we also train on SMDD and test on the MAD22 benchmark~\cite{synmad22}, which includes both landmark and GAN-based morphs. For completeness, we further report results on the MorDIFF~\cite{mordiff} extension, which provide challenging diffusion-based morphs built from the same FRLL identities. 
Unless stated otherwise, all results are obtained from the officially provided aligned images and reported using the same measures, enabling direct comparison with previously published results. Overall, this experimental design covers both cross-dataset and cross-morph generalization.

\section{Experiments and Results}
\label{sec:experiments}
\begin{table*}[ht]
\setlength{\tabcolsep}{4pt}
\centering
\caption{\textbf{Results comparison between our method and previous S-MAD solutions when tested on FRLL-Morph and trained on SMDD.} We report Equal Error Rate (EER \%) and BPCER (\%) at fixed APCER levels of 1\% and 20\% for different morphing attacks. Specific values unavailable in the original papers are marked with "-". The proposed method achieves the lowest average EER and demonstrates strong performance across different morphs.}
\label{tab:results_frll}
\resizebox{\linewidth}{!}{%
\begin{tabular}{@{}ccccccccccccccccccc@{}}
\toprule
\multirow{3}{*}[-1.5ex]{Method} & \multicolumn{3}{c}{OpenCV}                                                    & \multicolumn{3}{c}{FaceMorph}                                                                 & \multicolumn{3}{c}{WebMorph}                                                                  & \multicolumn{3}{c}{AMSL}                                                                           & \multicolumn{3}{c}{StyleGan2}                                                 & \multicolumn{3}{c}{Avg.}                                                                      \\ \cmidrule(l){2-19} 
                        & \multicolumn{1}{c}{\multirow{2}{*}{EER}} & \multicolumn{2}{c}{BPCER@APCER}           & \multicolumn{1}{c}{\multirow{2}{*}{EER}} & \multicolumn{2}{c}{BPCER@APCER}                           & \multicolumn{1}{c}{\multirow{2}{*}{EER}} & \multicolumn{2}{c}{BPCER@APCER}                           & \multicolumn{1}{c}{\multirow{2}{*}{EER}} & \multicolumn{2}{c}{BPCER@APCER}                           & \multicolumn{1}{c}{\multirow{2}{*}{EER}} & \multicolumn{2}{c}{BPCER@APCER}           & \multicolumn{1}{c}{\multirow{2}{*}{EER}} & \multicolumn{2}{c}{BPCER@APCER}                           \\ \cmidrule(lr){3-4} \cmidrule(lr){6-7} \cmidrule(lr){9-10} \cmidrule(lr){12-13} \cmidrule(lr){15-16} \cmidrule(l){18-19} 
                        & \multicolumn{1}{c}{}                     & \multicolumn{1}{c}{1\%}  & 20\% & \multicolumn{1}{c}{}                     & \multicolumn{1}{c}{1\%}          & 20\%         & \multicolumn{1}{c}{}                     & \multicolumn{1}{c}{1\%}          & 20\%         & \multicolumn{1}{c}{}                          & \multicolumn{1}{c}{1\%}          & 20\%         & \multicolumn{1}{c}{}                     & \multicolumn{1}{c}{1\%}  & 20\% & \multicolumn{1}{c}{}                     & \multicolumn{1}{c}{1\%}          & 20\%         \\ \midrule
OrthoMAD~\cite{orthomad}                & \multicolumn{1}{c}{\underline{0.73}}                 & \multicolumn{1}{c}{13.74} & 3.76  & \multicolumn{1}{c}{\underline{0.98}}                 & \multicolumn{1}{c}{0.98}          & 0.08          & \multicolumn{1}{c}{15.23}                & \multicolumn{1}{c}{15.23}         & 9.50          & \multicolumn{1}{c}{14.80}                     & \multicolumn{1}{c}{65.05}         & 10.89         & \multicolumn{1}{c}{{\underline{ 6.54}}}           & \multicolumn{1}{c}{13.74} & 3.76  & \multicolumn{1}{c}{7.66}                 & \multicolumn{1}{c}{21.75}         & 5.60          \\ 
IDistill~\cite{idistill}                & \multicolumn{1}{c}{2.46}                 & \multicolumn{1}{c}{6.14}  & 0.16  & \multicolumn{1}{c}{2.05}                 & \multicolumn{1}{c}{4.26}          & 0.16          & \multicolumn{1}{c}{4.01}                 & \multicolumn{1}{c}{14.41}         & 0.33          & \multicolumn{1}{c}{4.00}                      & \multicolumn{1}{c}{21.10}         & 2.85          & \multicolumn{1}{c}{\textbf{1.96}}        & \multicolumn{1}{c}{8.51}  & 0.08  & \multicolumn{1}{c}{\underline{2.90}}                 & \multicolumn{1}{c}{10.88}         & \textbf{0.72} \\ 
MixFaceNet-MAD~\cite{smdd_inception_mixfacenet}          & \multicolumn{1}{c}{4.39}                 & \multicolumn{1}{c}{26.47} & 1.47  & \multicolumn{1}{c}{3.87}                 & \multicolumn{1}{c}{23.53}         & 0.49          & \multicolumn{1}{c}{12.35}                & \multicolumn{1}{c}{80.39}         & 13.24         & \multicolumn{1}{c}{15.18}                     & \multicolumn{1}{c}{49.51}         & 11.76         & \multicolumn{1}{c}{8.99}                 & \multicolumn{1}{c}{42.16} & 4.41  & \multicolumn{1}{c}{8.96}                 & \multicolumn{1}{c}{44.41}         & 6.27          \\ 
PW-MAD~\cite{pwmad}                  & \multicolumn{1}{c}{2.42}                 & \multicolumn{1}{c}{22.06} & 0.49  & \multicolumn{1}{c}{2.20}                 & \multicolumn{1}{c}{26.47}         & 0.00          & \multicolumn{1}{c}{16.65}                & \multicolumn{1}{c}{80.39}         & 13.24         & \multicolumn{1}{c}{15.18}                     & \multicolumn{1}{c}{96.57}         & 5.88          & \multicolumn{1}{c}{16.64}                & \multicolumn{1}{c}{80.39} & 13.24 & \multicolumn{1}{c}{10.62}                & \multicolumn{1}{c}{61.18}         & 6.57          \\ 
WB-Avcivas~\cite{wb_avcivas}              & \multicolumn{1}{c}{7.91}                 & \multicolumn{1}{c}{-}     & -     & \multicolumn{1}{c}{17.11}                & \multicolumn{1}{c}{-}             & -             & \multicolumn{1}{c}{19.32}                & \multicolumn{1}{c}{-}             & -             & \multicolumn{1}{c}{18.23}                     & \multicolumn{1}{c}{-}             & -             & \multicolumn{1}{c}{14.87}                & \multicolumn{1}{c}{-}     & -     & \multicolumn{1}{c}{15.49}                & \multicolumn{1}{c}{-}             & -             \\ 
Inception-MAD~\cite{smdd_inception_mixfacenet}           & \multicolumn{1}{c}{5.38}                 & \multicolumn{1}{c}{38.73} & 0.98  & \multicolumn{1}{c}{3.17}                 & \multicolumn{1}{c}{30.39}         & 0.49          & \multicolumn{1}{c}{9.86}                 & \multicolumn{1}{c}{53.92}         & 2.94          & \multicolumn{1}{c}{10.79}                     & \multicolumn{1}{c}{72.06}         & 4.90          & \multicolumn{1}{c}{11.37}                & \multicolumn{1}{c}{72.06} & 6.86  & \multicolumn{1}{c}{8.11}                 & \multicolumn{1}{c}{53.43}         & 3.23          \\ 
MADation (ViT-B)~\cite{caldeira2025madation}        & \multicolumn{1}{c}{2.97}                 & \multicolumn{1}{c}{4.41}  & 0.49  & \multicolumn{1}{c}{1.35}                 & \multicolumn{1}{c}{1.47}          & 0.00          & \multicolumn{1}{c}{\underline{3.42}}                 & \multicolumn{1}{c}{5.88}          & 0.00          & \multicolumn{1}{c}{\underline{3.85}}                      & \multicolumn{1}{c}{12.07}         & 2.41          & \multicolumn{1}{c}{17.21}                & \multicolumn{1}{c}{54.50} & 13.10 & \multicolumn{1}{c}{5.76}                 & \multicolumn{1}{c}{15.67}         & {\underline{ 3.20}}    \\ \midrule
Ours                    & \multicolumn{1}{c}{\textbf{0.00}}        & \multicolumn{1}{c}{\textbf{0.00}}  & \textbf{0.00}  & \multicolumn{1}{c}{\textbf{0.00}}        & \multicolumn{1}{c}{\textbf{0.00}} & \textbf{0.00} & \multicolumn{1}{c}{\textbf{0.00}}        & \multicolumn{1}{c}{\textbf{0.00}} & \textbf{0.00} & \multicolumn{1}{c}{\textbf{0.00}}             & \multicolumn{1}{c}{\textbf{0.00}} & \textbf{0.00} & \multicolumn{1}{c}{9.25}                 & \multicolumn{1}{c}{24.02} & 2.45  & \multicolumn{1}{c}{\textbf{1.85}}        & \multicolumn{1}{c}{\textbf{4.80}} & 4.19          \\ \bottomrule
\end{tabular}%
}
\end{table*}
In this section, we analyze the performance of our method across different benchmark datasets. We compare our method against state-of-the-art supervised methods and highlight its robust performance. We also provide ablations to quantify the contribution of global and local Fourier features.

\vspace{0.5em}
\noindent\textbf{Implementation Details.}
We conduct the experiments on NVIDIA GeForce RTX 5090 with 32GB of VRAM, and the framework utilized was Pytorch~\cite{pytorch}. To ensure consistency across experiments, Fourier features were extracted using images of uniform size 500x500, as previous works~\cite {pca_morph, selfmad++} suggest that higher-resolution images typically preserve more morphing artifacts. For Fourier feature extraction, we use a batch size of 128, and for the global features, a support vector machine (SVM) with an RBF kernel was trained. In order to obtain the semantic face regions, we use the RETINAFace~\cite{retinaface} implementation, which provides 106 facial landmarks per face. For each region $r$, we select a fixed subset of landmark indices and crop a bounding box around the corresponding points with a small margin. Regarding the pairwise interactions in the MRF, we use a fixed parameter $\beta = 0.9$, tuned on SMDD and held constant across evaluations, and set the final fusion weight to $\lambda = 0.6$. Regarding the dataset proportion of positive and negative samples, we use the same number of bona fide images and morphs for the training phase.

\begin{table*}[ht]
\setlength{\tabcolsep}{4pt}
\centering
\caption{\textbf{Results comparison between our method and previous S-MAD solutions when tested on MAD22~\cite{synmad22} and trained on SMDD~\cite{smdd_inception_mixfacenet}.} Specific values unavailable in the original papers are marked with "-". While D-FW-CDCN~\cite{dfw} achieves the lowest average EER, our approach achieves the second-best average EER without relying on heavy 3D-aware multi-task architectures.}
\label{tab:results_mad22}
\resizebox{\linewidth}{!}{%
\begin{tabular}{@{}cccccccccccccccccccccc@{}}
\toprule
\multirow{3}{*}[-1.5ex]{Method} & \multicolumn{3}{c}{OpenCV} & \multicolumn{3}{c}{FaceMorph} & \multicolumn{3}{c}{WebMorph} & \multicolumn{3}{c}{MIPGAN-I} & \multicolumn{3}{c}{MIPGAN-II} & \multicolumn{3}{c}{MorDIFF} & \multicolumn{3}{c}{Avg.\textsuperscript{\textdagger}} \\ \cmidrule(l){2-22} 
& \multicolumn{1}{c}{\multirow{2}{*}{EER}} & \multicolumn{2}{c}{BPCER@APCER}
& \multicolumn{1}{c}{\multirow{2}{*}{EER}} & \multicolumn{2}{c}{BPCER@APCER}
& \multicolumn{1}{c}{\multirow{2}{*}{EER}} & \multicolumn{2}{c}{BPCER@APCER}
& \multicolumn{1}{c}{\multirow{2}{*}{EER}} & \multicolumn{2}{c}{BPCER@APCER}
& \multicolumn{1}{c}{\multirow{2}{*}{EER}} & \multicolumn{2}{c}{BPCER@APCER}
& \multicolumn{1}{c}{\multirow{2}{*}{EER}} & \multicolumn{2}{c}{BPCER@APCER}
& \multicolumn{1}{c}{\multirow{2}{*}{EER}} & \multicolumn{2}{c}{BPCER@APCER}
\\ \cmidrule(lr){3-4} \cmidrule(lr){6-7} \cmidrule(lr){9-10} \cmidrule(lr){12-13} \cmidrule(lr){15-16} \cmidrule(lr){18-19} \cmidrule(l){21-22}
& \multicolumn{1}{c}{} & \multicolumn{1}{c}{1\%} & 20\%
& \multicolumn{1}{c}{} & \multicolumn{1}{c}{1\%} & 20\%
& \multicolumn{1}{c}{} & \multicolumn{1}{c}{1\%} & 20\%
& \multicolumn{1}{c}{} & \multicolumn{1}{c}{1\%} & 20\%
& \multicolumn{1}{c}{} & \multicolumn{1}{c}{1\%} & 20\%
& \multicolumn{1}{c}{} & \multicolumn{1}{c}{1\%} & 20\%
& \multicolumn{1}{c}{} & \multicolumn{1}{c}{1\%} & 20\%
\\ \midrule
MixFaceNet-MAD~\cite{smdd_inception_mixfacenet}
& \multicolumn{1}{c}{8.33} & \multicolumn{1}{c}{44.31} & 1.96
& \multicolumn{1}{c}{4.60} & \multicolumn{1}{c}{55.88} & 0.00
& \multicolumn{1}{c}{18.20} & \multicolumn{1}{c}{92.86} & 15.20
& \multicolumn{1}{c}{16.70} & \multicolumn{1}{c}{89.21} & 11.76
& \multicolumn{1}{c}{20.62} & \multicolumn{1}{c}{91.64} & 20.10
& \multicolumn{1}{c}{8.50} & \multicolumn{1}{c}{62.67} & 2.94
& \multicolumn{1}{c}{13.69} & \multicolumn{1}{c}{74.78} & 9.80
\\
Inception-MAD~\cite{smdd_inception_mixfacenet}
& \multicolumn{1}{c}{7.52} & \multicolumn{1}{c}{-} & -
& \multicolumn{1}{c}{0.00} & \multicolumn{1}{c}{-} & -
& \multicolumn{1}{c}{18.00} & \multicolumn{1}{c}{-} & -
& \multicolumn{1}{c}{10.90} & \multicolumn{1}{c}{-} & -
& \multicolumn{1}{c}{16.22} & \multicolumn{1}{c}{-} & -
& \multicolumn{1}{c}{5.30} & \multicolumn{1}{c}{-} & -
& \multicolumn{1}{c}{10.53} & \multicolumn{1}{c}{-} & \textbf{-}
\\
MorphHRNet~\cite{synmad22}
& \multicolumn{1}{c}{5.69} & \multicolumn{1}{c}{33.82} & 1.47
& \multicolumn{1}{c}{5.90} & \multicolumn{1}{c}{48.04} & 1.47
& \multicolumn{1}{c}{{\underline{ 9.80}}} & \multicolumn{1}{c}{{\underline{ 56.86}}} & \textbf{3.92}
& \multicolumn{1}{c}{15.30} & \multicolumn{1}{c}{75.98} & 11.27
& \multicolumn{1}{c}{10.41} & \multicolumn{1}{c}{61.27} & 2.94
& \multicolumn{1}{c}{-} & \multicolumn{1}{c}{-} & -
& \multicolumn{1}{c}{9.42} & \multicolumn{1}{c}{55.19} & \textbf{4.21}
\\
Con-Text Net A~\cite{synmad22}
& \multicolumn{1}{c}{17.48} & \multicolumn{1}{c}{74.02} & 16.18
& \multicolumn{1}{c}{0.00} & \multicolumn{1}{c}{0.00} & 0.00
& \multicolumn{1}{c}{26.20} & \multicolumn{1}{c}{93.14} & 31.86
& \multicolumn{1}{c}{12.30} & \multicolumn{1}{c}{59.31} & 6.37
& \multicolumn{1}{c}{12.91} & \multicolumn{1}{c}{\textbf{59.31}} & 5.88
& \multicolumn{1}{c}{-} & \multicolumn{1}{c}{-} & -
& \multicolumn{1}{c}{13.78} & \multicolumn{1}{c}{57.16} & 12.06
\\
D-FW-MixFaceNet~\cite{dfw}
& \multicolumn{1}{c}{13.72} & \multicolumn{1}{c}{-} & -
& \multicolumn{1}{c}{{\underline{ 0.10}}} & \multicolumn{1}{c}{-} & -
& \multicolumn{1}{c}{10.80} & \multicolumn{1}{c}{-} & -
& \multicolumn{1}{c}{{\underline{ 6.70}}} & \multicolumn{1}{c}{-} & -
& \multicolumn{1}{c}{\textbf{6.61}} & \multicolumn{1}{c}{-} & -
& \multicolumn{1}{c}{-} & \multicolumn{1}{c}{-} & -
& \multicolumn{1}{c}{7.59} & \multicolumn{1}{c}{-} & -
\\
D-FW-CDCN~\cite{dfw}
& \multicolumn{1}{c}{\textbf{0.30}} & \multicolumn{1}{c}{-} & -
& \multicolumn{1}{c}{0.00} & \multicolumn{1}{c}{-} & -
& \multicolumn{1}{c}{\textbf{0.00}} & \multicolumn{1}{c}{-} & -
& \multicolumn{1}{c}{11.90} & \multicolumn{1}{c}{-} & -
& \multicolumn{1}{c}{14.11} & \multicolumn{1}{c}{-} & -
& \multicolumn{1}{c}{-} & \multicolumn{1}{c}{-} & -
& \multicolumn{1}{c}{\textbf{5.26}} & \multicolumn{1}{c}{-} & -
\\
MADation (ViT-B)~\cite{caldeira2025madation}
& \multicolumn{1}{c}{3.85} & \multicolumn{1}{c}{23.53} & \textbf{0.00}
& \multicolumn{1}{c}{0.00} & \multicolumn{1}{c}{0.00} & 0.00
& \multicolumn{1}{c}{10.80} & \multicolumn{1}{c}{\textbf{51.47}} & {\underline{ 4.41}}
& \multicolumn{1}{c}{33.37} & \multicolumn{1}{c}{94.12} & 52.94
& \multicolumn{1}{c}{22.21} & \multicolumn{1}{c}{84.80} & 26.47
& \multicolumn{1}{c}{\underline{1.10}} & \multicolumn{1}{c}{1.94} & 0
& \multicolumn{1}{c}{14.05} & \multicolumn{1}{c}{50.78} & {16.76}
\\ \midrule
Ours
& \multicolumn{1}{c}{{\underline{ 2.54}}} & \multicolumn{1}{c}{\textbf{7.35}} & {\underline{ 1.47}}
& \multicolumn{1}{c}{\textbf{0.00}} & \multicolumn{1}{c}{\textbf{0.00}} & \textbf{0.00}
& \multicolumn{1}{c}{18.40} & \multicolumn{1}{c}{86.76} & 17.16
& \multicolumn{1}{c}{\textbf{2.45}} & \multicolumn{1}{c}{\textbf{6.37}} & \textbf{0.98}
& \multicolumn{1}{c}{{\underline{ 7.21}}} & \multicolumn{1}{c}{100.00} & \textbf{1.47}
& \multicolumn{1}{c}{\textbf{0.00}} & \multicolumn{1}{c}{\textbf{0.00}} & \textbf{0.00}
& \multicolumn{1}{c}{\underline{6.12}} & \multicolumn{1}{c}{\textbf{40.10}} & {\underline{ 4.22}}
\\ \bottomrule
\end{tabular}%
}
\noindent\parbox{\linewidth}{\vspace*{0.5em} \footnotesize\textsuperscript{\textdagger} Avg. is calculated across all morphing attacks excluding MorDIFF due to unavailable values which would lead to unfair comparisons.}
\end{table*}

\subsection{Baseline Comparison}
We compare our method with previous S-MAD works on both evaluation protocols.

Table~\ref{tab:results_frll} reports results for the first protocol.  Across all attacks, our approach achieves the lowest average EER ($1.85\%$) and clearly improves over strong supervised baselines such as IDistill~\cite{idistill} and MADation (ViT-B)~\cite{caldeira2025madation}. The gains are particularly pronounced for classical warping-based morphs (OpenCV, FaceMorph, WebMorph, AMSL), where our method achieves almost perfect detection (EER below $0.2\%$ and \mbox{BPCER=$0\%$} at both operating points). This behaviour is consistent with how these morph sets are constructed and with prior observations on landmark-based morphing pipelines. FRLL-Morphs~\cite{frll_morph} builds upon the AMSL protocols and provides predefined morph pairs together with classical landmark-based morphing tools, which are known to introduce characteristic artefacts, such as ghosting and local blending inconsistencies. As a result, these attacks are comparatively easier for S-MAD methods to exploit, even though they remain highly effective at attacking face recognition systems. This interpretation is further supported by our ablation study in Tables~\ref{tab:ablation_study1_frll} and \ref{tab:ablation_study1_mad22}. When using only the local branch, the EER on FRLL-Morphs~\cite{frll_morph} drops to 0\% for all landmark-based attacks, while remaining higher for the StyleGAN2 attack. GAN-based morphs tend to suppress local artefacts, making them inherently more challenging to separate from bona fide images, which explains the gap in performance between the landmark-based and StyleGAN2 attacks. In contrast, IDistill~\cite{idistill} explicitly targets identity entanglement by estimating whether two identities are present in the image (using $id_1 \cdot id_2$), making it less dependent on low-level artefacts and therefore particularly effective for StyleGAN2 morphs. Nevertheless, our global–local Fourier model is competitive and still yields lower average EER and BPCER than deep learning baselines, despite using only lightweight spectral descriptors rather than full CNN backbones.

In contrast, the MAD22~\cite{synmad22} benchmark is explicitly designed to yield challenging and realistic morphs. For each FRLL attack, the authors first obtain the embeddings with a strong face recognition model and then select only the most similar identity pairs as candidates for morphing. These pairs are then morphed using multiple pipelines, including both landmark-based approaches and high-quality GAN-based methods, which substantially reduce visible artefacts while preserving the identity blend. This protocol construction results in a more difficult detection scenario, explaining the higher error rates on MAD22~\cite{synmad22} presented in Table~\ref{tab:results_mad22} compared to FRLL-Morphs~\cite{frll_morph}. Beyond the original MAD22 attacks, we additionally report results on MorDIFF~\cite{mordiff}, and, since diffusion-based morph attacks are recent, some earlier baselines do not report MorDIFF results; therefore, MorDIFF is excluded from the average to avoid unfair comparisons, but we deliberately include it to assess robustness against modern and more realistic morph generation pipelines. Table~\ref{tab:results_mad22} shows that the D-FW~\cite{dfw} variant, which relies on a multi-task framework with additional 3D facial cues, obtains the best average EER. Despite this, our approach ranks second in terms of average EER ($6.12\%$ vs.\ $5.26\%$ for D-FW-CDCN~\cite{dfw}), achieving a BPCER@APCER=1\% of $40.10\%$ and BPCER@APCER=20\% of $4.22\%$.

In particular, the proposed method is highly effective on the FaceMorph, MIPGAN-I/II and MorDIFF~\cite{mordiff} attacks, while the WebMorph attack remains the most difficult across methods. Our method achieves perfect separation on MorDIFF~\cite{mordiff}, outperforming MADation~\cite{caldeira2025madation}, the strongest competing approach with available results. Beyond quantitative results, Figure~\ref{fig:tsne_visualization} shows that diffusion-based morphs form a compact and well-separated cluster away from bona fide samples in the feature space, indicating that the proposed method capture distinctive artefacts introduced by the diffusion process. These results demonstrate that our approach not only generalizes across classical attacks, but also remains effective against emerging diffusion-based morphing techniques.
It is worth mentioning that our model achieves this performance without any auxiliary 3D supervision or heavy backbones, highlighting that Fourier-domain residuals are a competitive alternative for S-MAD under strict computational constraints. Although 3D-aware multi-task architectures such as D-FW~\cite{dfw} achieve slightly lower EERs, their higher complexity results in a significant increase in computational demands, making them less suitable for resource-constrained deployments. Moreover, integrating such cues into our framework is an interesting direction for future work, but currently beyond the scope of this paper.

\subsection{Ablation Studies}
In order to evaluate the impact of the isolated global and local features, we report results that show the impact of each component.

\vspace{0.5em}
\noindent\textbf{Impact of global and local features.} Tables~\ref{tab:ablation_study1_frll} and~\ref{tab:ablation_study1_mad22} analyze the contribution of global Fourier residuals, local region-wise features, and their fusion.

\begin{table}[t]
\setlength{\tabcolsep}{4pt}
\centering
\caption{\textbf{Impact of global and local Fourier residual features on S-MAD performance.} EER (\%) is reported, and the fusion of global and local features improves S-MAD performance, confirming the complementarity of global and regional cues.}
\label{tab:ablation_study1_frll}
\resizebox{\linewidth}{!}{%
\begin{tabular}{@{}cccc@{}}
\toprule
\multicolumn{1}{c}{Tested on FRLL-Morph~\cite{frll_morph}}      & \multicolumn{1}{c}{\makecell{Global \\Features (G)}} & \multicolumn{1}{c}{\makecell{Local \\Features (L)}} & G + L          \\ \midrule 
\multicolumn{1}{c}{FRLL-OpenCV}    & \multicolumn{1}{c}{0.98}            & \multicolumn{1}{c}{0.00}           & \textbf{0.00} \\ 
\multicolumn{1}{c}{FRLL-FaceMorph} & \multicolumn{1}{c}{0.49}            & \multicolumn{1}{c}{0.00}           & \textbf{0.00} \\ 
\multicolumn{1}{c}{FRLL-WebMorph}  & \multicolumn{1}{c}{0.41}            & \multicolumn{1}{c}{0.08}           & \textbf{0.00} \\ 
\multicolumn{1}{c}{FRLL-AMSL}      & \multicolumn{1}{c}{2.16}            & \multicolumn{1}{c}{0.00}           & \textbf{0.00} \\ 
\multicolumn{1}{c}{FRLL-StyleGan2} & \multicolumn{1}{c}{{\textbf{ 6.37}}}      & \multicolumn{1}{c}{18.09}          & 9.25          \\ \midrule
\multicolumn{1}{c}{Average}        & \multicolumn{1}{c}{2.08}            & \multicolumn{1}{c}{3.63}           & \textbf{1.85} \\ \bottomrule
\end{tabular}%
}
\end{table}

On the first protocol (Table~\ref{tab:ablation_study1_frll}), global features alone already provide strong performance (average EER $2.08\%$), while local features alone are particularly effective for most attacks but struggle on StyleGAN2 (average EER $3.63\%$ dominated by this case). Fusing global and local information consistently improves the average EER to $1.85\%$, and either matches or outperforms the best single component on four out of five attacks. This confirms that global and local features capture complementary cues: global features are more robust to GAN-based morphs, whereas local residuals excel at detecting classical morphing pipelines.

On MAD22~\cite{synmad22} (Table~\ref{tab:ablation_study1_mad22}), the benefit of fusion is even more evident.  Global and local features alone yield similar but relatively high average EERs ($16.62\%$ and $16.65\%$), with each component specializing in different attacks (e.g.\ global features handle MIPGAN-I well, while local features are better for WebMorph and MIPGAN-II). The fused model reduces the average EER to $5.10\%$, substantially outperforming both individual streams and achieving close-to-best performance for every attack. From this ablation, we conclude that combining general and region-wise Fourier residuals within the MRF is essential for robustness across diverse morphing mechanisms.

\begin{table}[t]
\setlength{\tabcolsep}{4pt}
\centering
\caption{\textbf{Impact of global and local Fourier residual features on S-MAD performance.} EER (\%) is reported, and the fusion of global with local features reduces the average EER, demonstrating that combining holistic and structured regional evidence helps achieve robust generalization across diverse morphing attacks.}
\label{tab:ablation_study1_mad22}
\resizebox{.9\linewidth}{!}{%
\begin{tabular}{@{}cccc@{}}
\toprule
\multicolumn{1}{c}{Tested on MAD22~\cite{synmad22}}      & \multicolumn{1}{c}{\makecell{Global\\Features (G)}} & \multicolumn{1}{c}{\makecell{Local\\Features (L)}} & G + L          \\ \midrule
\multicolumn{1}{c}{OpenCV}    & \multicolumn{1}{c}{10.37}           & \multicolumn{1}{c}{{ 6.37}}    & {\textbf{ 2.54}}    \\ 
\multicolumn{1}{c}{FaceMorph} & \multicolumn{1}{c}{9.20}            & \multicolumn{1}{c}{0.00}           & \textbf{0.00} \\ 
\multicolumn{1}{c}{WebMorph}  & \multicolumn{1}{c}{45.80}           & \multicolumn{1}{c}{21.57} & \textbf{18.40 }        \\ 
\multicolumn{1}{c}{MIPGAN-I}      & \multicolumn{1}{c}{\textbf{1.30}}   & \multicolumn{1}{c}{58.70}          & 2.45 \\ 
\multicolumn{1}{c}{MIPGAN-II} & \multicolumn{1}{c}{33.03}           & \multicolumn{1}{c}{{13.24}}    & {\textbf{ 7.21}}    \\ 
\multicolumn{1}{c}{MorDIFF} & \multicolumn{1}{c}{0.00}           & \multicolumn{1}{c}{{0.00}}    & {\textbf{0.00}} \\ \midrule
\multicolumn{1}{c}{Average}        & \multicolumn{1}{c}{16.62}           & \multicolumn{1}{c}{{16.65}}    & {\textbf{ 5.10}}    \\ \bottomrule

\end{tabular}%
}
\end{table}

\subsection{Feature Visualization}
\begin{figure}[ht]
\centering
\begin{minipage}[c]{0.47\linewidth}
  \centering
  \includegraphics[width=\linewidth]{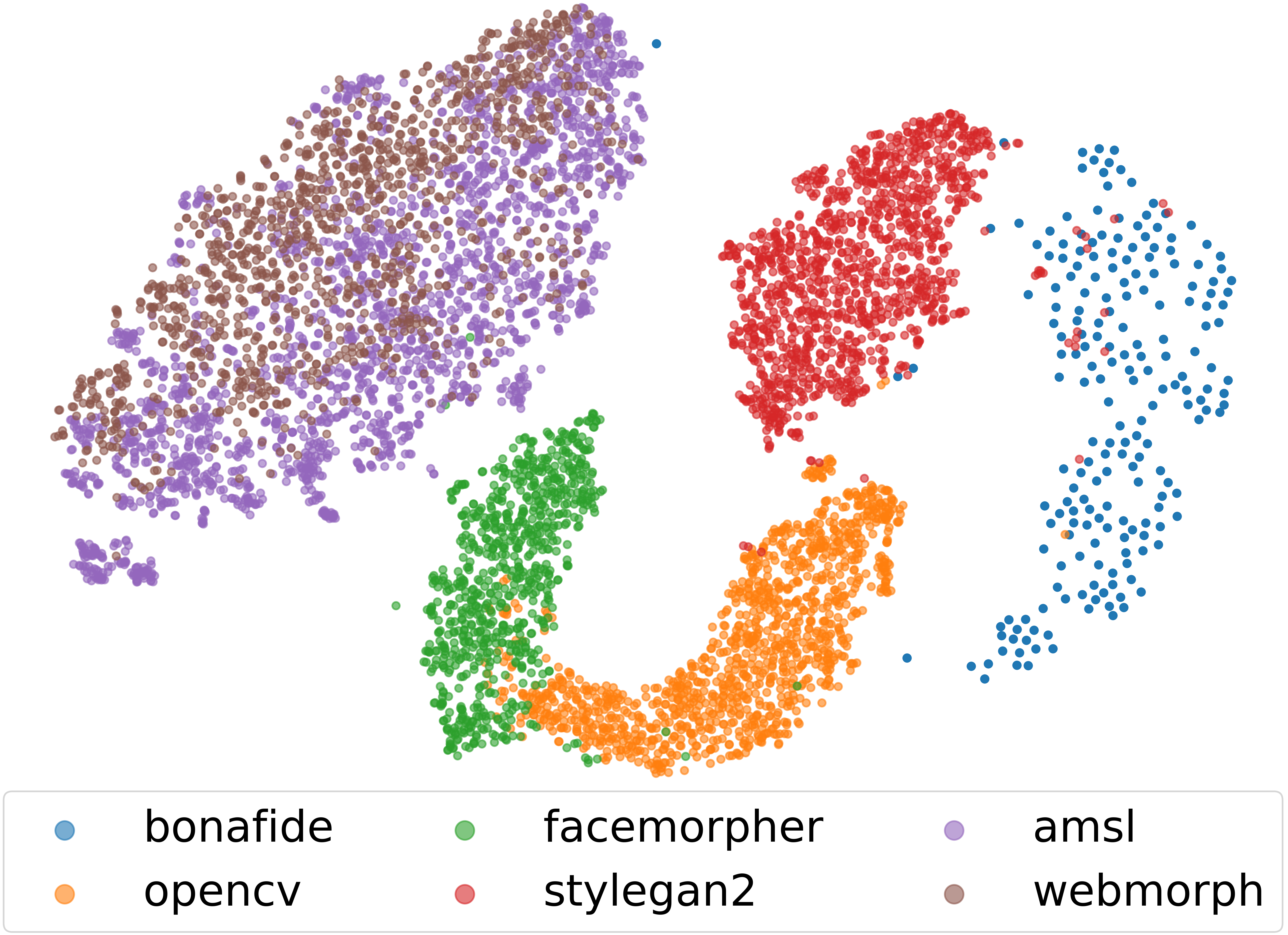}
\end{minipage}
\hspace{0.4em}\vrule width 0.5pt\hspace{0.4em}
\begin{minipage}[c]{0.47\linewidth}
  \centering
  \vspace*{0.30cm} 
  \includegraphics[width=\linewidth]{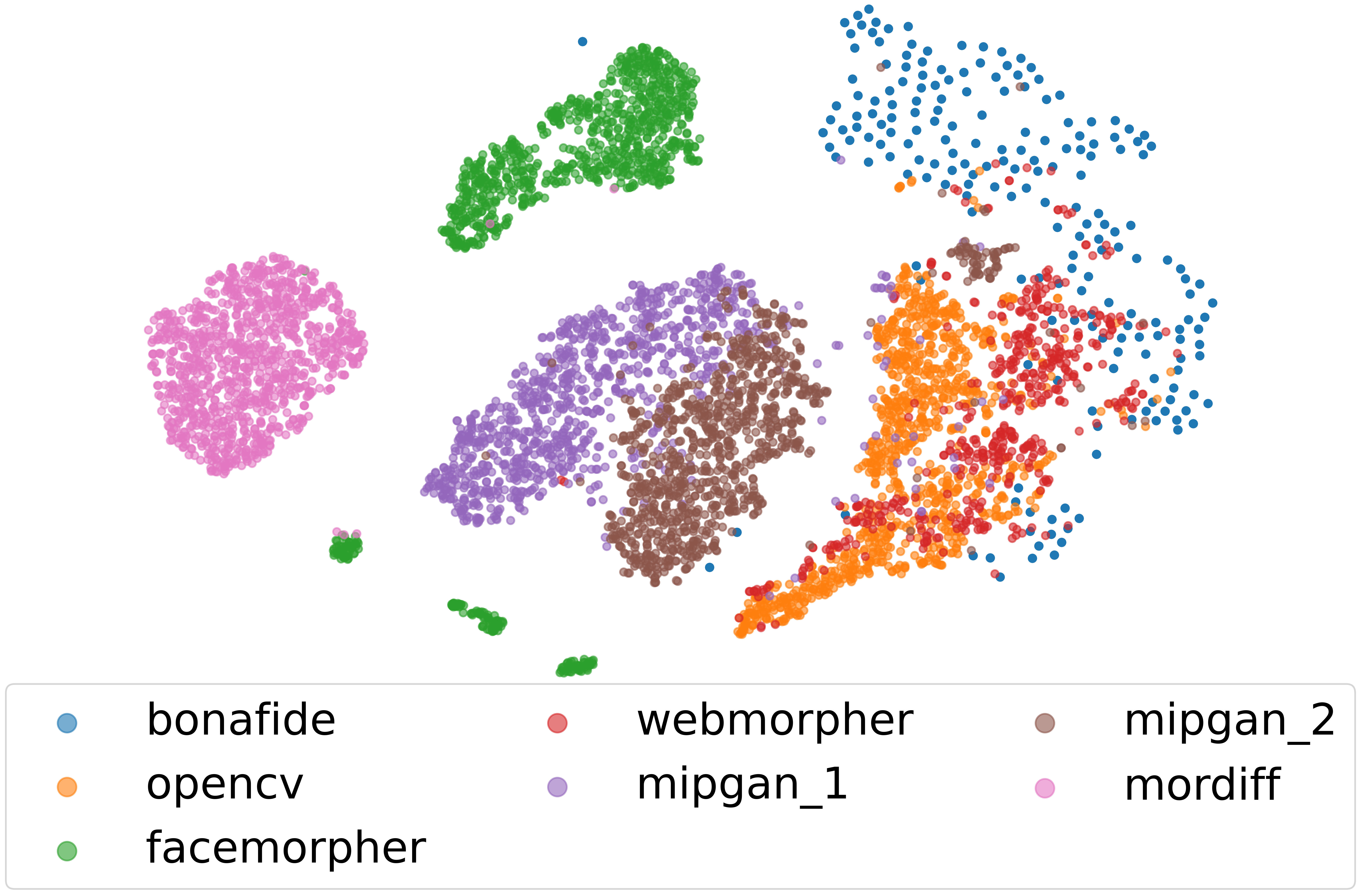}
\end{minipage}
\caption{\textbf{t-SNE visualization of frequency-domain global features obtained from different morphing attacks of FRLL-Morph~\cite{frll_morph} (left) and MAD22~\cite{synmad22} (right).} For MAD22, the WebMorph attack follows a similar distribution, making it a hard one.}
\label{fig:tsne_visualization}
\end{figure}

To better understand the behaviour of the Fourier-domain features, we visualize them using t-SNE. Figure~\ref{fig:tsne_visualization} shows the t-SNE embeddings of FRLL-Morph~\cite{frll_morph} and MAD22~\cite{synmad22} global features. The different morphing attacks form well-separated clusters, and bona fide samples occupy a distinct region of the space. Regarding the FRLL-Morph~\cite{frll_morph} dataset, the AMSL and WebMorph attacks partially overlap, indicating that these pipelines produce similar spectral artefacts, which explains why they are harder to distinguish from each other. Nevertheless, bona fide samples remain clearly separated from all morph clusters, supporting the high detection rates obtained on this dataset.

As for the MAD22~\cite{synmad22} dataset, the distributions of some attacks, particularly WebMorph, overlap more with bona fide samples, revealing a more challenging decision boundary. This overlap correlates with the higher EERs observed for these attacks and highlights MAD22 as a significantly harder benchmark. Overall, the visualizations confirm that the proposed frequency-domain residuals encode discriminative, attack-sensitive structure while still enabling a compact and interpretable representation for S-MAD.

\section{Conclusion}
This work introduced a novel frequency-domain method for detecting single-image face morphing attacks (S-MAD). Our approach combines global Fourier residual features with region-based spectral features, using a Markov Random Field. By removing the power-law natural trend and thereby obtaining residual features, our approach differs from traditional deep learning approaches, remains interpretable and reduces reliance on dataset-specific biases that hamper generalization across different morphing techniques. Our experiments were carried out on FRLL-Morph and MAD22, and show that combining global and local features helps our method generalize well from the SMDD training data to new morphing attacks. On FRLL-Morph, our approach outperforms or matches recent supervised S-MAD methods, and it performs competitively on MAD22, even though our models are much simpler than multi-task or 3D-aware ones. Then, the ablation studies show that global and local Fourier residuals provide complementary discriminative cues, and that using an MRF to maintain facial region consistency is important for handling both landmark-based and GAN-based morphs.
As future work, we plan to expand our framework by adding new cues, such as 3D shape information, within the same frequency-domain and MRF setup, making the method even more robust to new morph types. Furthermore, we could expand it by exploring contrastive learning on Fourier residual embeddings to strengthen bona fide clusters and increase separation from morph samples, potentially improving robustness against challenging attack types that produce less consistently localized artefacts. Our spectral residual modeling might also work well for other biometric systems or attack-detection tasks, especially when computational resources are limited.

{\small
\bibliographystyle{ieee}
\bibliography{egbib}
}

\end{document}